\documentclass[10pt,journal,compsoc]{IEEEtran}

\ifCLASSOPTIONcompsoc
  \usepackage{cite}
\else
  \usepackage{cite}
\fi

\ifCLASSINFOpdf
\else
\fi

\usepackage{graphicx}
\usepackage{comment}
\usepackage{amsmath,amssymb} 
\allowdisplaybreaks[1]
\usepackage{color}
\usepackage{booktabs}
\usepackage{array}
\usepackage[table]{xcolor}
\usepackage[caption=false,font=footnotesize]{subfig}
\usepackage{diagbox}
\usepackage{multirow}
\usepackage{hyperref}
\usepackage{relsize}

\newcommand{\figref}[1]{Fig.~\ref{fig:#1}}
\newcommand{\tblref}[1]{Table~\ref{tbl:#1}}
\newcommand{\secref}[1]{Section~\ref{sec:#1}}

\DeclareMathOperator*{\argmin}{arg\,min}

\definecolor{cobalt}{rgb}{0.0, 0.28, 0.67}
\hypersetup{
colorlinks=true,
urlcolor=black,
linkcolor=cobalt,
citecolor=cobalt}

\newcommand{\setup}{
\begin{figure}[t]
 \centering
  \vspace{-.1in}
  \includegraphics[width=0.99\linewidth]{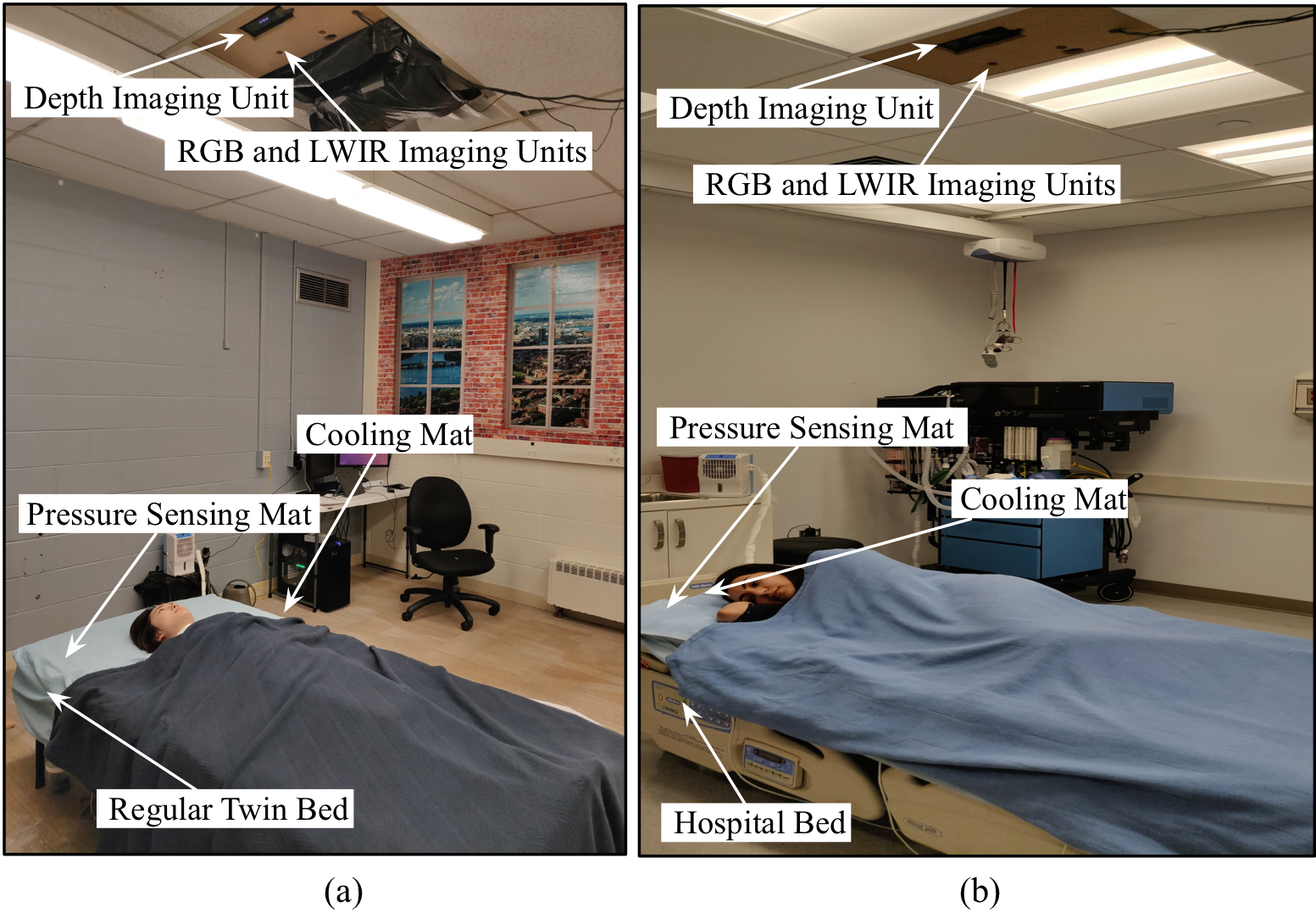}
  \caption{Our Multimodal in-bed pose data collection setup, (a) in a regular bedroom, (b) in a simulated hospital room.}
        \vspace{-.1in}
\label{fig:setup}
    \vspace{-.15in}
\end{figure}
}
\newcommand{\multimodal}{
\begin{figure}[t]
    \centering
    \includegraphics[width=0.99\linewidth]{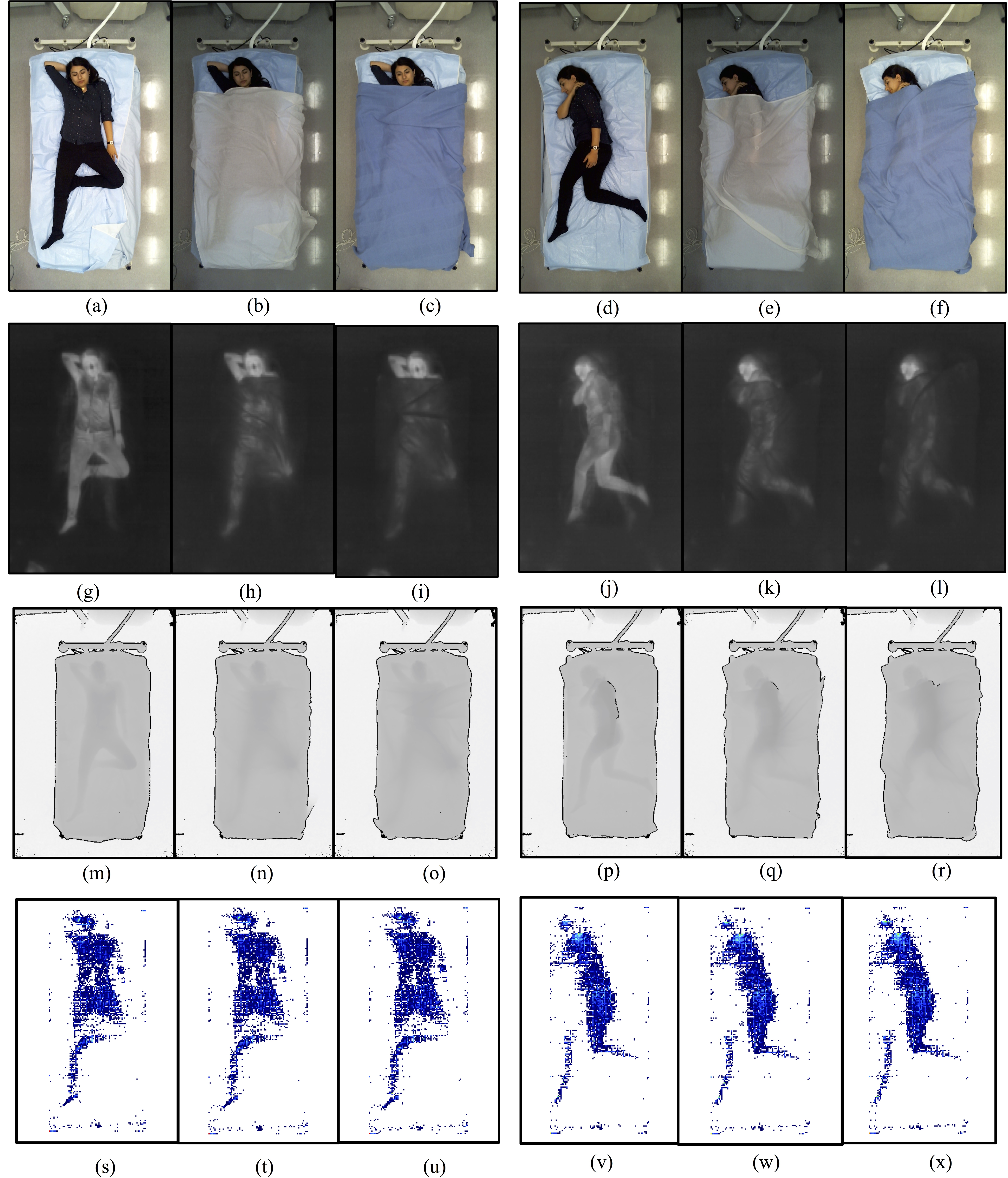}
    \caption{SLP image data samples from in-bed supine and side postures: (a-f) show images captured using an RGB webcam,  (g-l) show images captured using an LWIR camera, (m-r) shows images captured using a depth camera, and (s-x) shows images captured using a pressure mat. These images are taken from the participants without cover and with two different types (one thin and one thick) of covers.}
    \label{fig:multimodal}
    \vspace{-.2in}
\end{figure}
}
\newcommand{\figTempAmbi}{
\begin{figure}[t]
 \centering
  \vspace{-.1in}
  \subfloat{\label{fig:errorLabel}\includegraphics[width=0.99\linewidth,{trim=0in 0in 0in 0in,
  clip=true}]{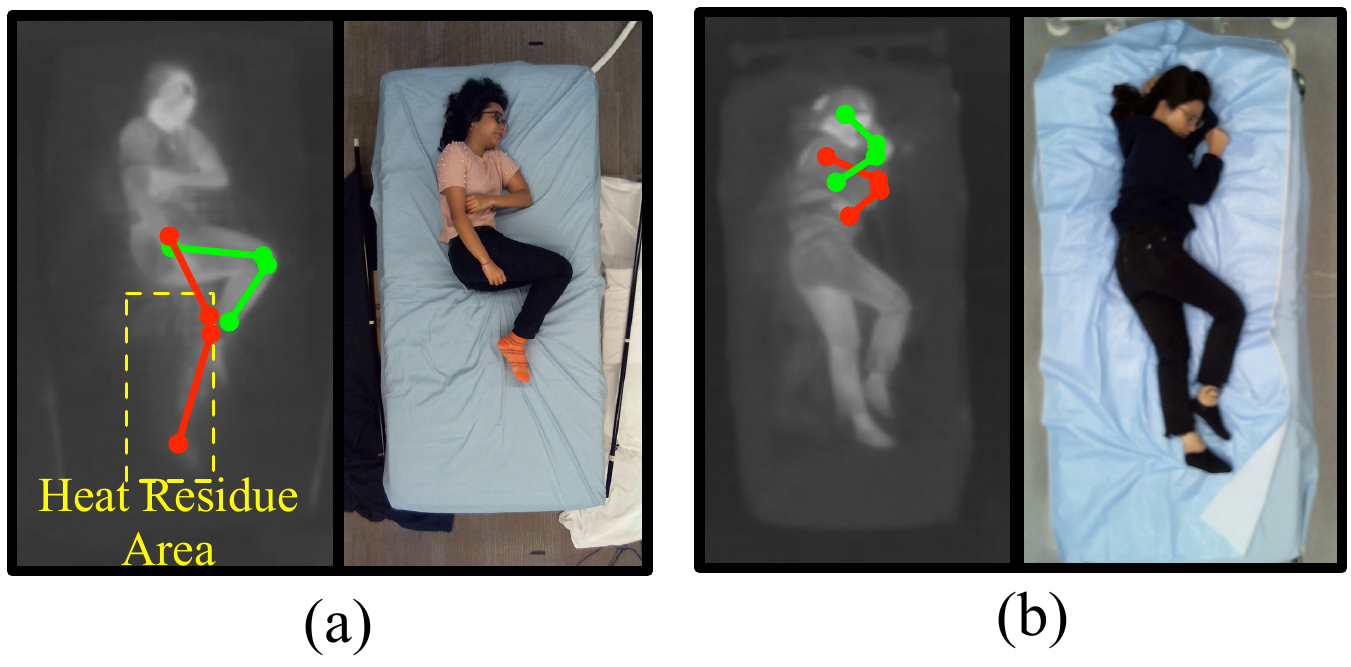}}
  \caption{Pose ambiguities in LWIR images with their corresponding RGB images, (a) false leg pose (in red) caused by the heat residue in the LWIR image,(b) false arm pose (in red) due to the cuddled limbs. The correct limb poses are given in green.}
        \vspace{-.1in}
\label{fig:tempAmbi}
    \vspace{-.1in}
\end{figure}
}
\newcommand{\figHomoErr}{
\begin{figure}[t]
    \vspace{-.1in}
    \centering
    \includegraphics[width=0.8\linewidth]{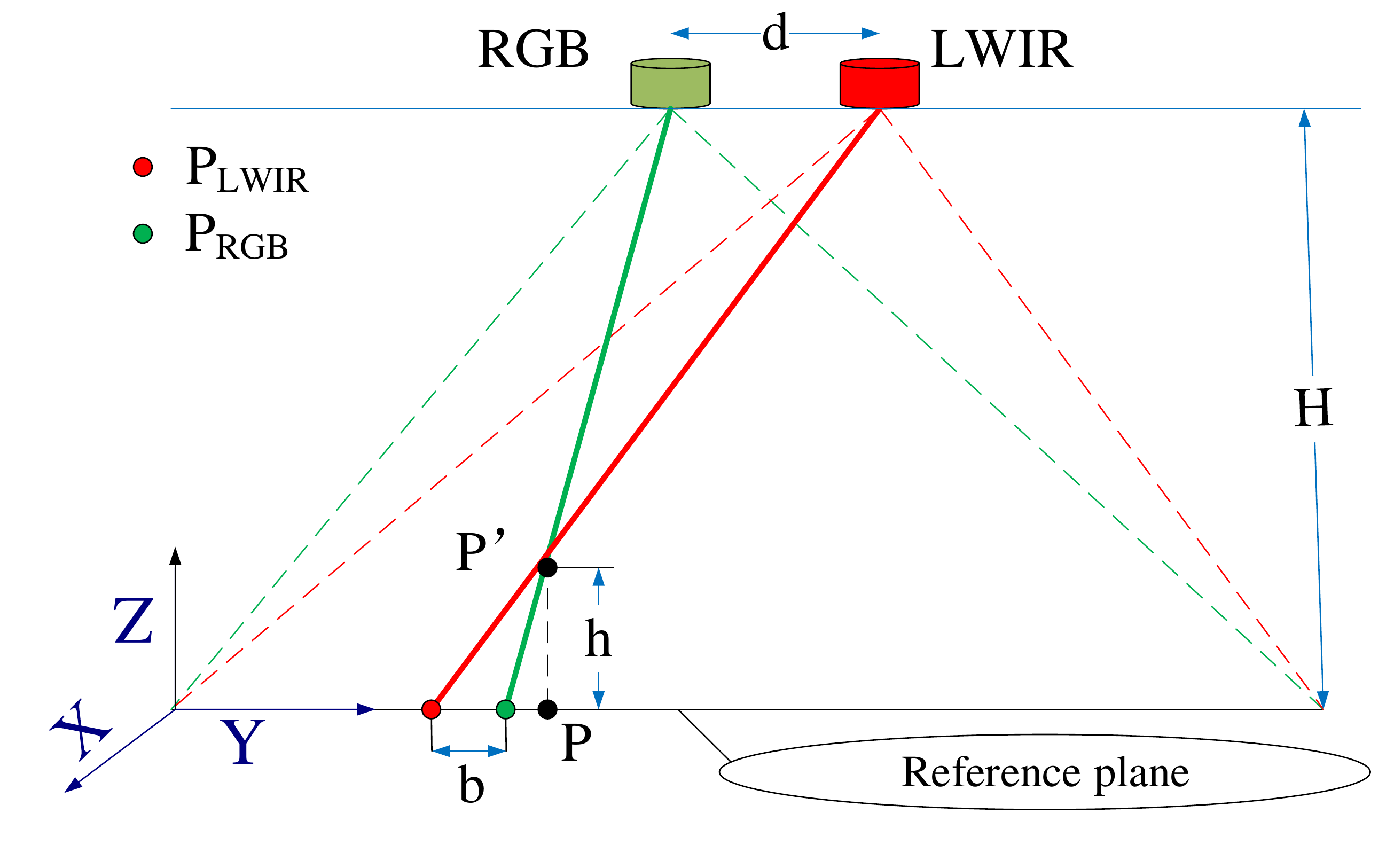}
    \caption{Ghosting error and between-camera mapping bias caused by elevation from reference plane with a working example between RGB and LWIR imaging modalities.}
    \label{fig:homoErr}
    \vspace{-.2in}
\end{figure}
}
\newcommand{\skelsPM}{
\begin{figure}[t]
 \centering
  \subfloat{\label{fig:setupSim}\includegraphics[width=0.9\linewidth,{trim=0in 0in 0in 0in,
  clip=true}]{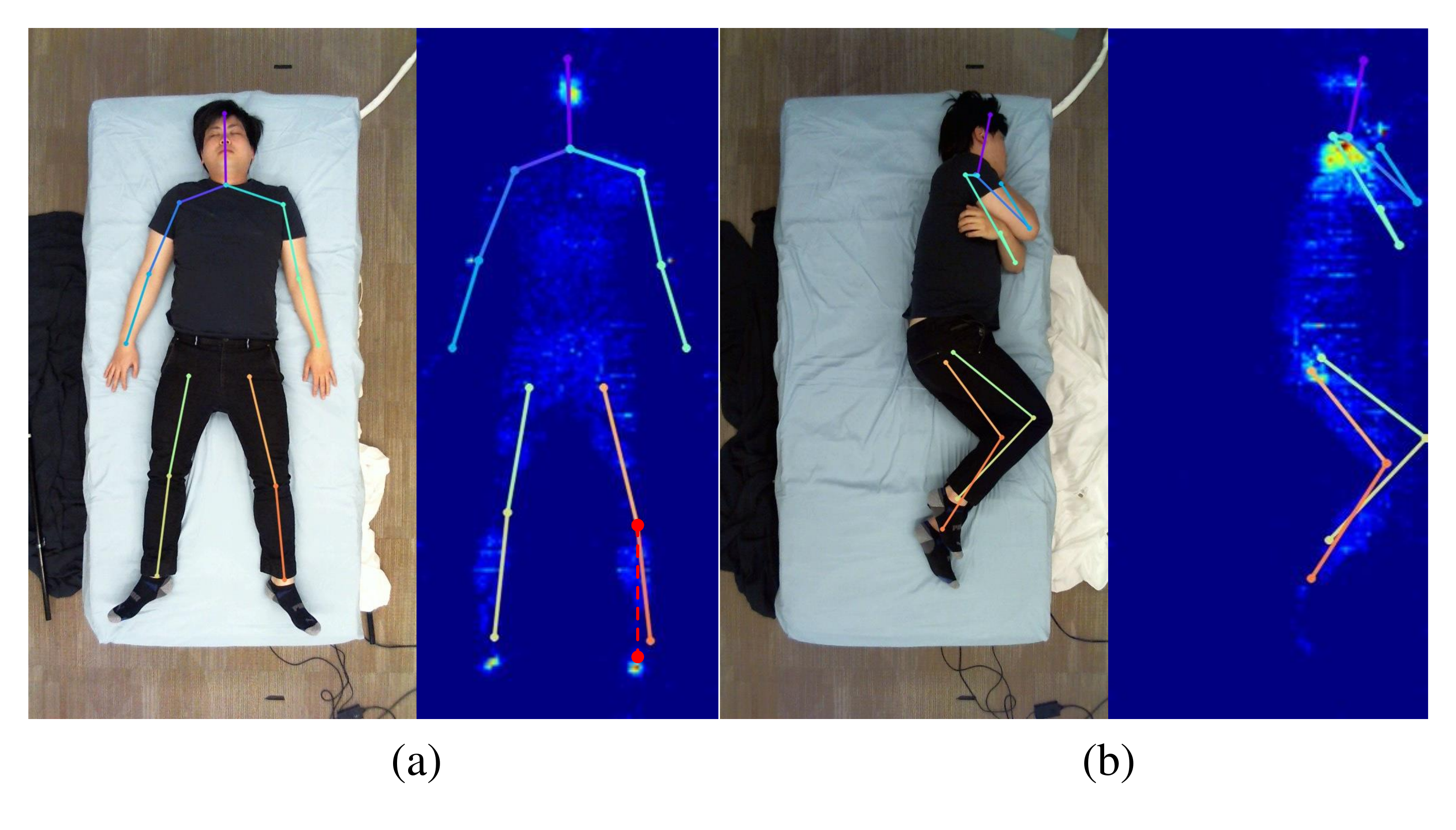}}
  \caption{Demos of PM ground truth generation via physical hyperparameter tuning (PHPT) of guideline III in: (a) a supine pose, and (b) a right lying pose. Red dash line shows direct annotation, intuitively.}
        \vspace{-.1in}
\label{fig:skelsPM}
    \vspace{-.15in}
\end{figure}
}
\newcommand{\alignModule}{
\begin{figure}[t]
 \centering
 \includegraphics[width=.8\linewidth]{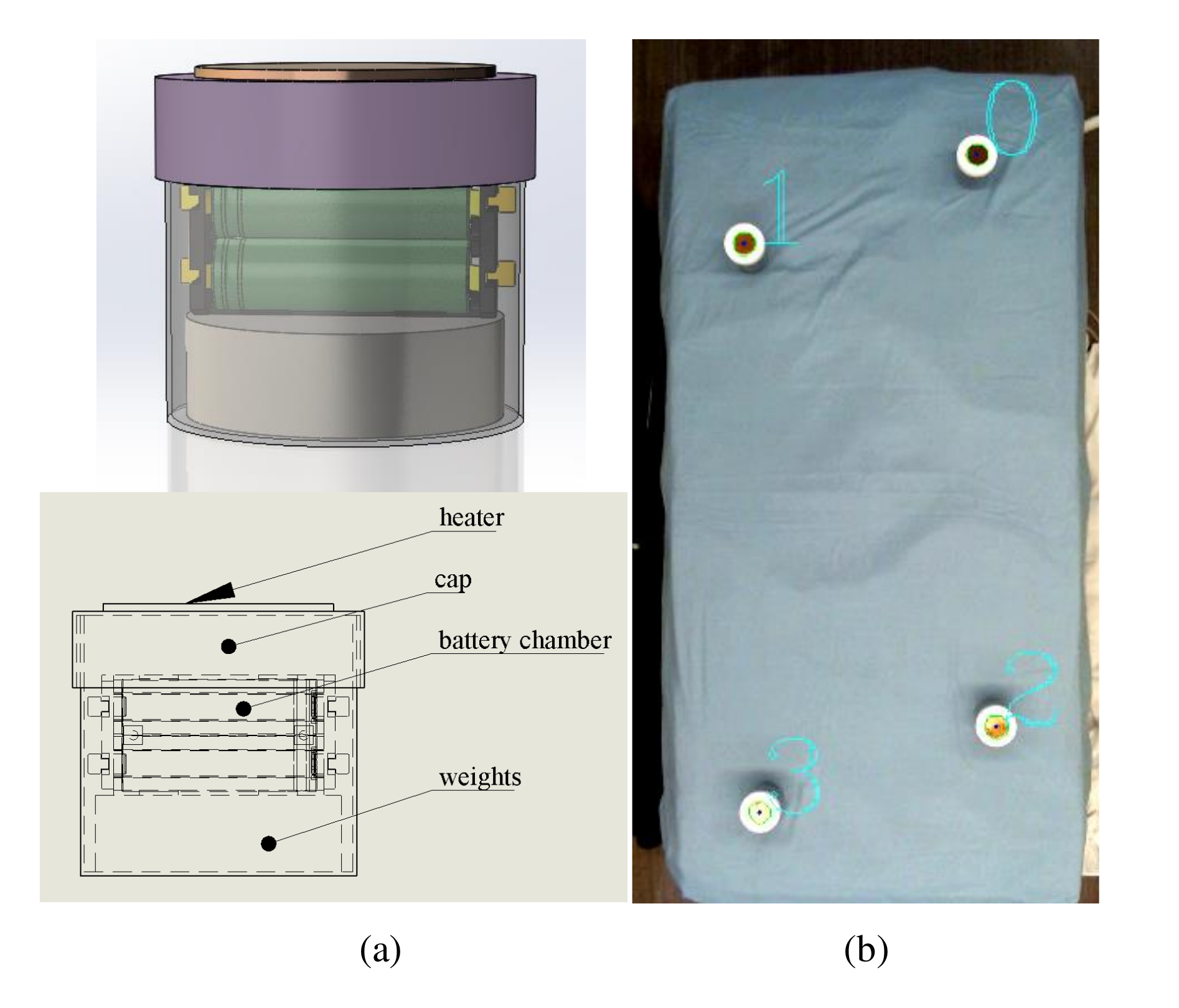}
   \caption{SLP cross domain alignment design: (a) alignment markers design, and (b) automatic center extraction in RGB imaging.}
\label{fig:alignModule}
\vspace{-.2in}
\end{figure}
}

\newcommand{\phyDist}{
\begin{figure*}[h]
 \centering
  \subfloat[]{\label{fig:height}\includegraphics[width=0.33\linewidth,{trim=0in 0in 0in 0in,
  clip=true}]{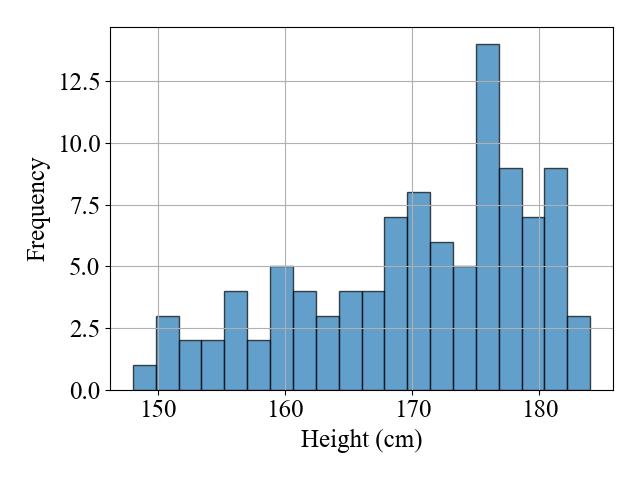}}
   \subfloat[]{\label{fig:weight}\includegraphics[width=0.33\linewidth,{trim=0in 0in 0in 0in,
  clip=true}]{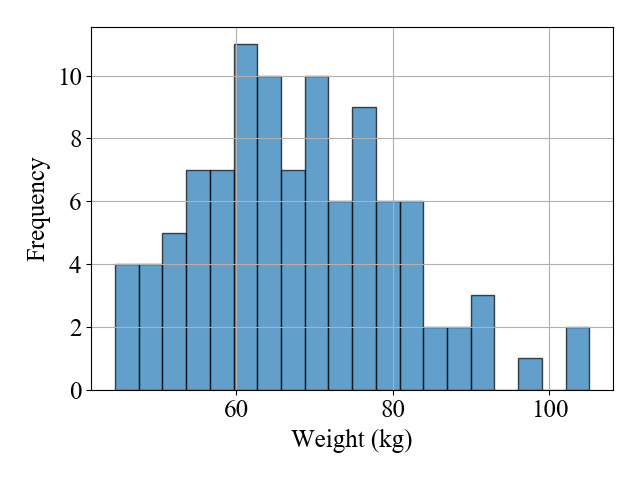}}
 \subfloat[]{\label{fig:tailor}\includegraphics[width=0.33\linewidth,{trim=0in 0in 0in 0in,
  clip=true}]{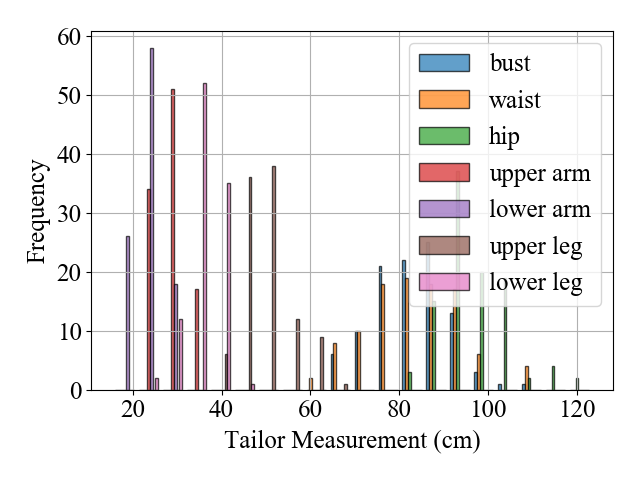}}
  \caption{Distribution of the measured person-specific parameters: (a) height (cm), (b) weight (kg), (c) tailor measurements (cm).}
\label{fig:phyDist}
\end{figure*}
}
\newcommand{\gtDist}{
\begin{figure}[t]
 \centering
 \subfloat[]{\label{fig:ankle_distWor}\includegraphics[width=0.45\linewidth,{trim=0in 0in 0in 0in,
  clip=true}]{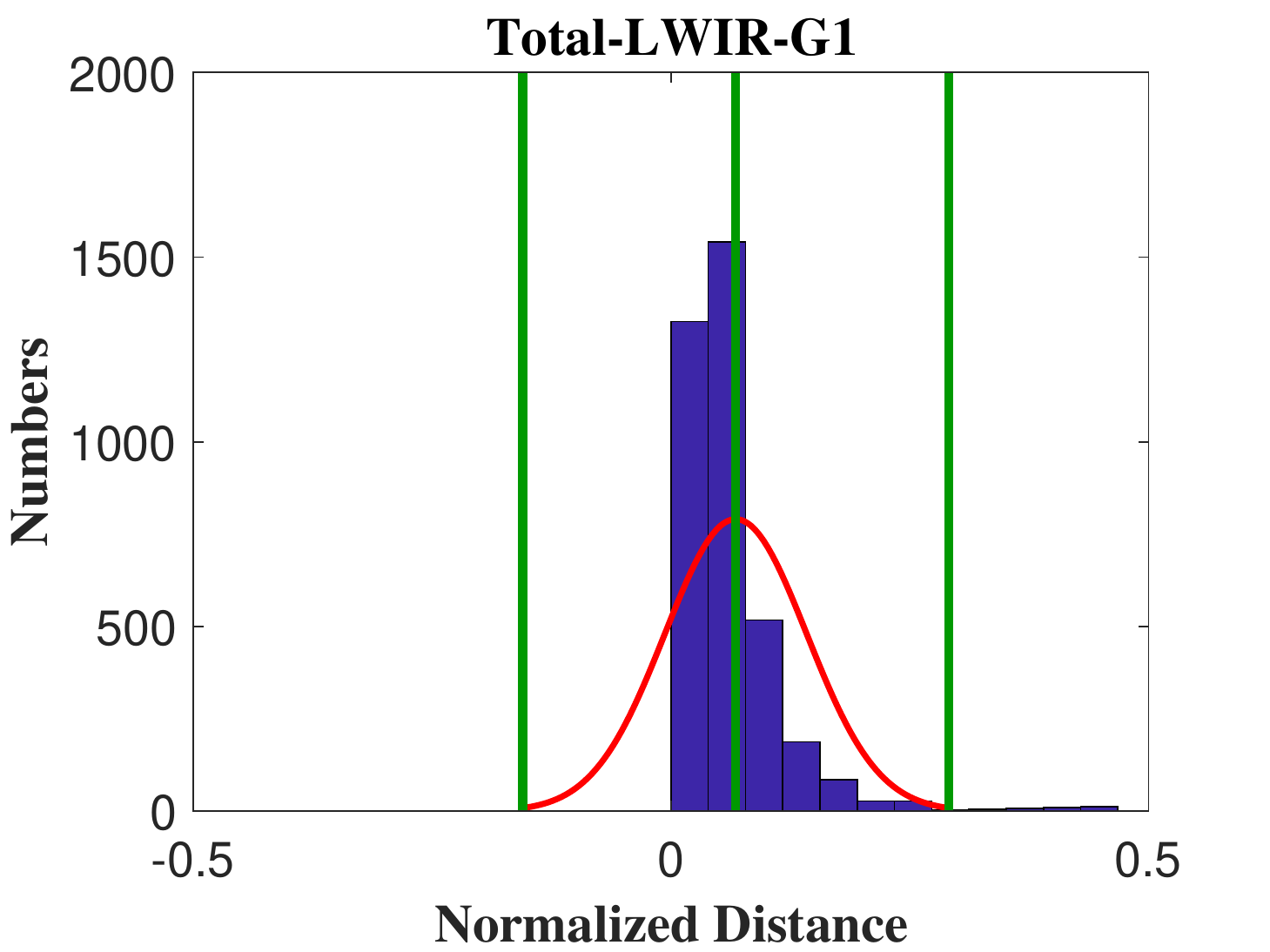}}
  \subfloat[]{\label{fig:knee_distWor}\includegraphics[width=0.45\linewidth,{trim=0in 0in 0in 0in,
  clip=true}]{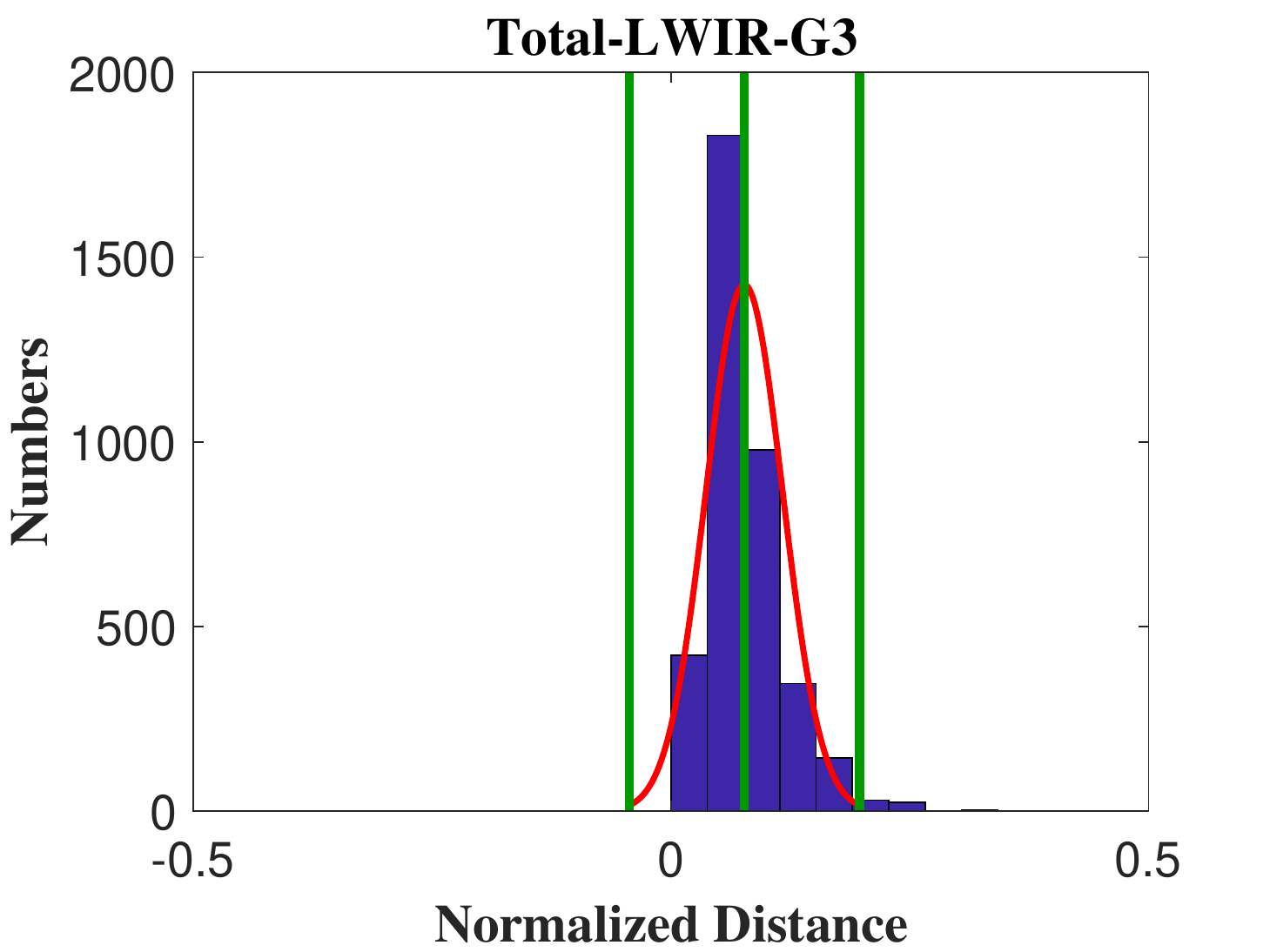}}
   \caption{Truncated histogram of normalized distance from the gold standard labels (using LWIR-G123) for labels generated using: (a) LWIR-G1, and (b) LWIR-G3. A Gaussian curve is fitted with green vertical lines as the mean and 3 standard deviation bounds.}
\label{fig:gtDist}
\vspace{-.2in}
\end{figure}
}
\newcommand{\pckDana}{
\begin{figure*}[t]
 \centering
 \subfloat{\includegraphics[width=1\linewidth,{trim=0in 0in 0in 0in,
  clip=true}]{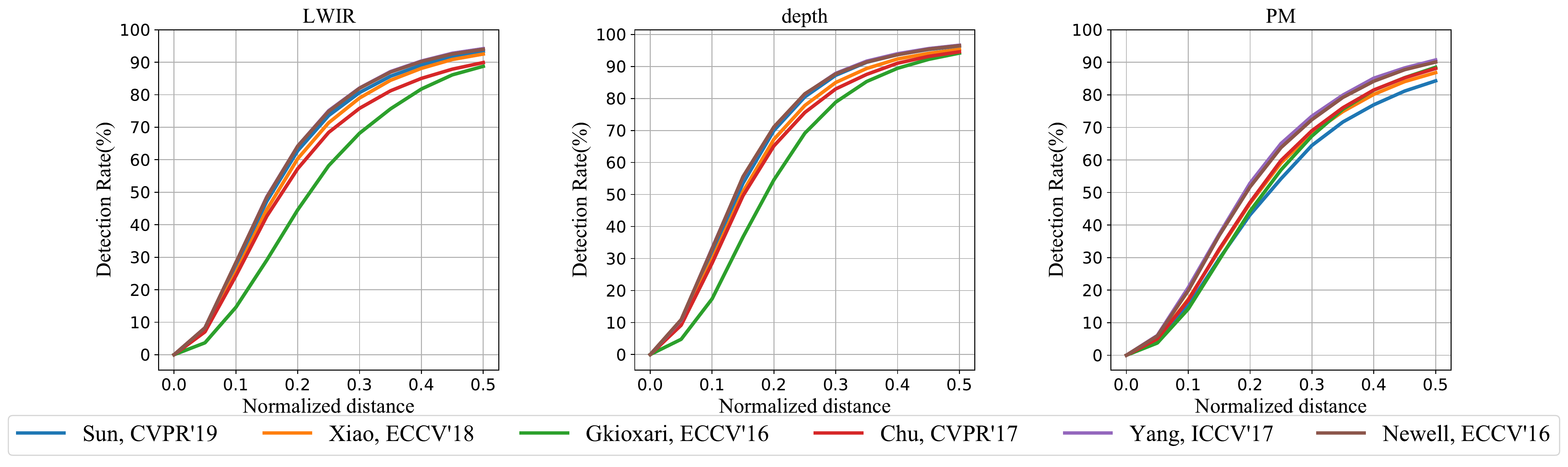}}
   \caption{PCKh pose estimation performance of state-of-the-arts on LWIR, depth, and PM modalities under home setting}
\label{fig:pckDana}
\end{figure*}
}
\newcommand{\demoTwo}{
\begin{figure*}[t]
\vspace{-.11in}
 \centering
 \subfloat{\includegraphics[width=1\linewidth,{trim=0in 0in 0in 0in,
  clip=true}]{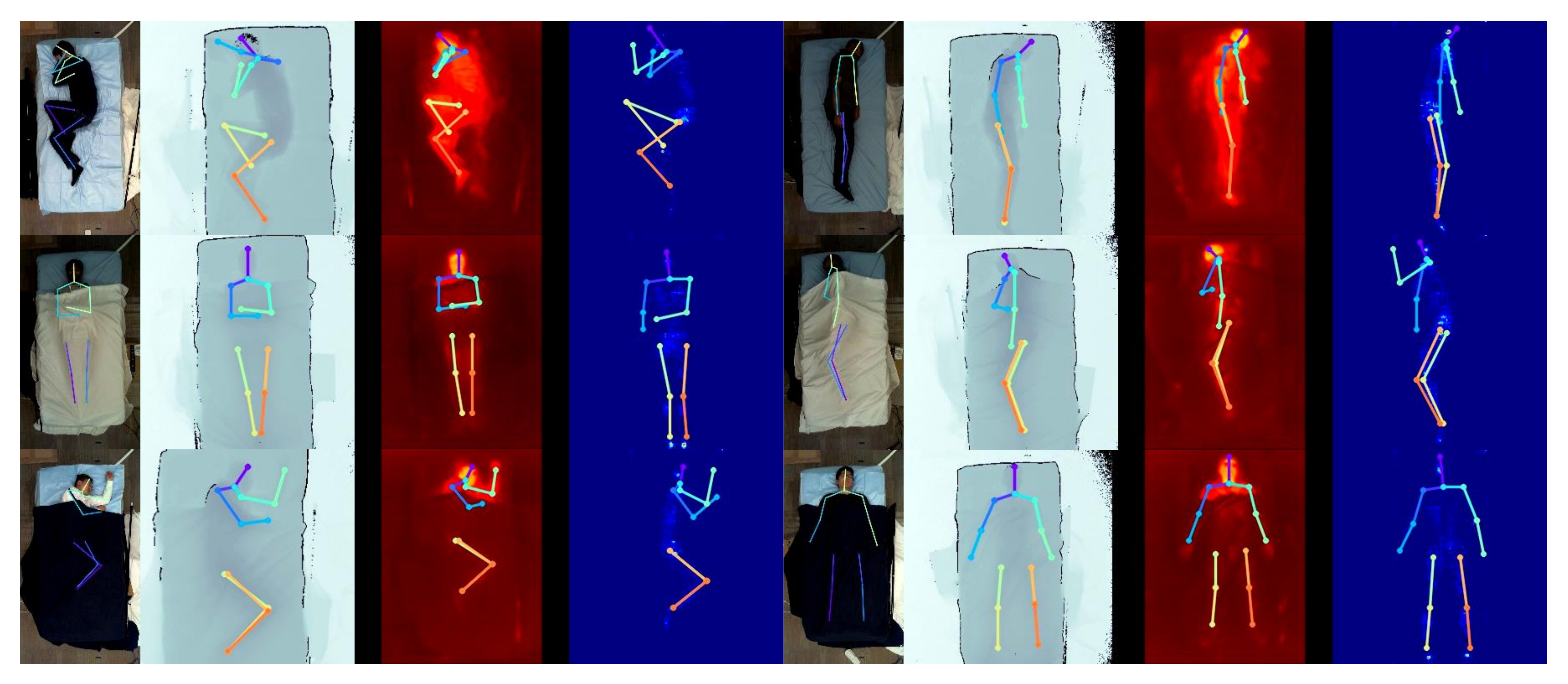}}
   \caption{Some qualitative results of in-bed  pose estimation based on Sun, CVPR'19\cite{sun2019deep}. RGB images are given with the ground truth pose along side the inference results from depth, LWIR, and PM modalities.}
\label{fig:demoTwo}
\vspace{-.2in}
\end{figure*}
}
\newcommand{\pckSim}{
\begin{figure}[t]
\vspace{-.11in}
 \centering
 \subfloat{\includegraphics[width=1\linewidth,{trim=0in 0in 0in 0in,
  clip=true}]{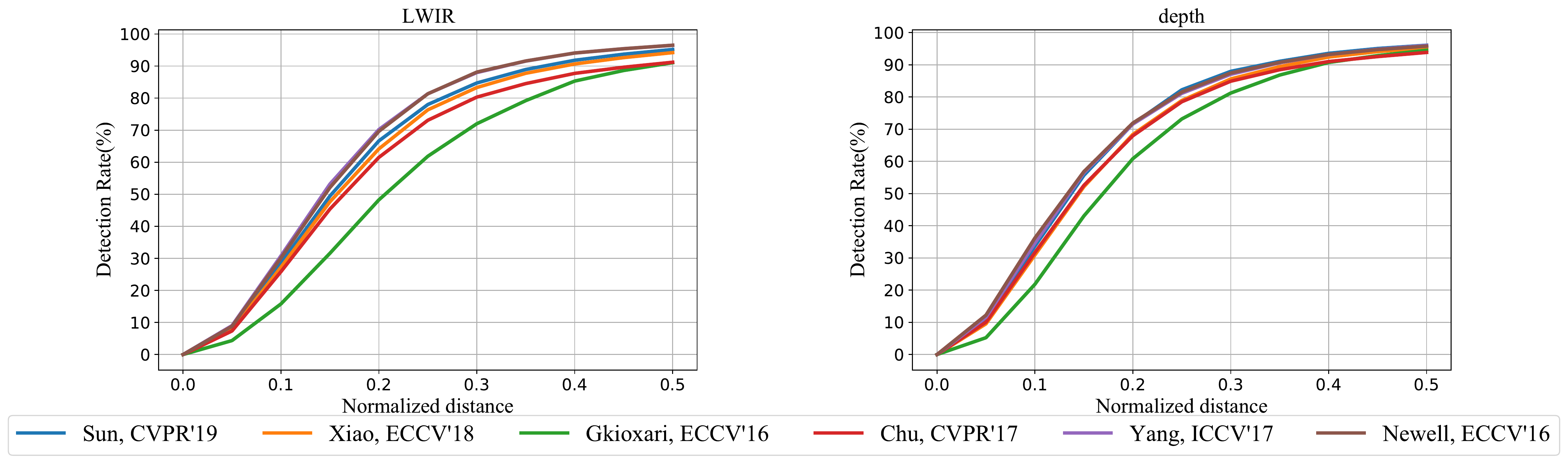}}
   \caption{PCKh pose estimation performance of state-of-the-arts on LWIR and depth modalities under hospital setting.}
\label{fig:pckSim}
\vspace{-.2in}
\end{figure}
}
\newcommand{\pckDanaJts}{
\begin{figure*}[t]
\vspace{-.11in}
 \centering
 \subfloat{\includegraphics[width=1\linewidth,{trim=0in 0in 0in 0in,
  clip=true}]{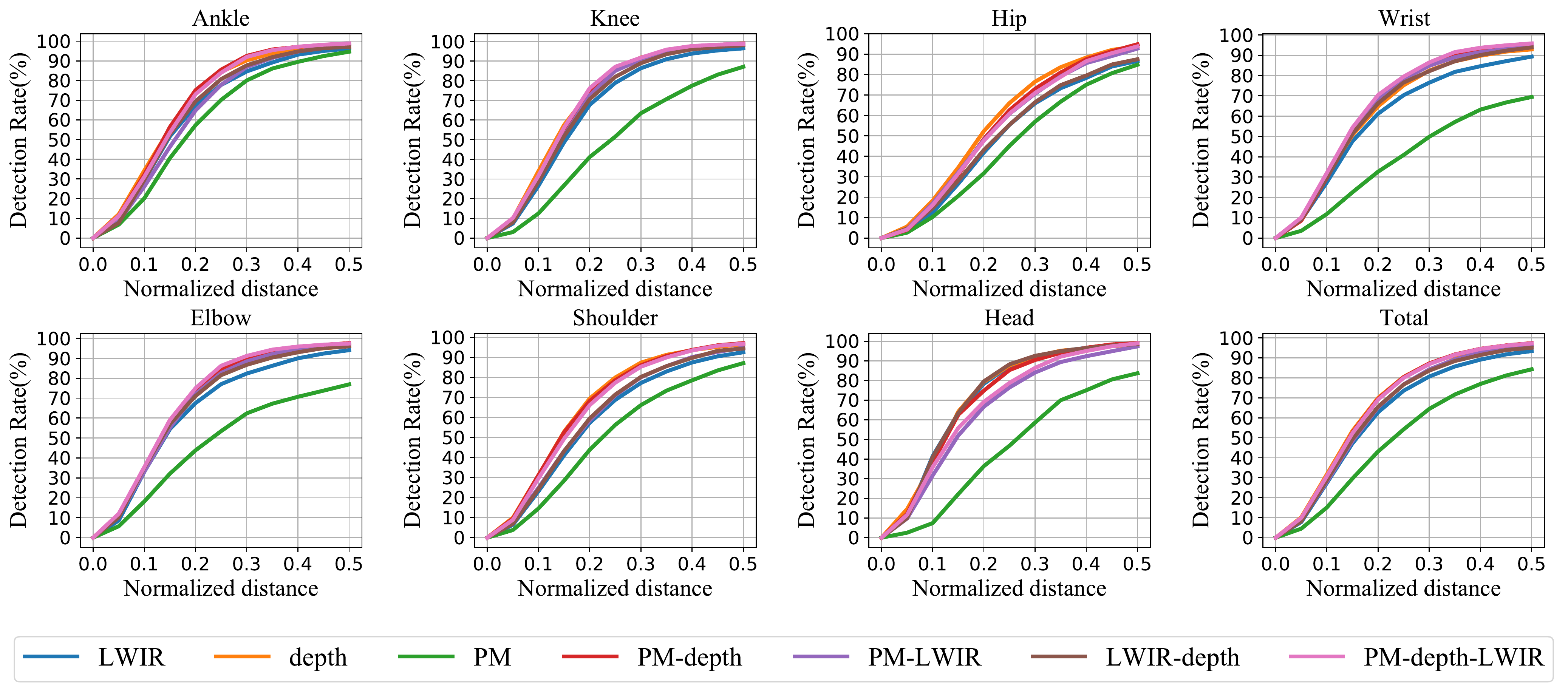}}
   \caption{PCKh pose estimation performance of \cite{sun2019deep}  using images in LWIR, depth, and PM modality and their combinations.}
\label{fig:pckDanaJts}
\vspace{-.2in}
\end{figure*}
}
\newcommand{\fuseVis}{
\begin{figure}[h]
    \centering
    \includegraphics[width=0.95\linewidth]{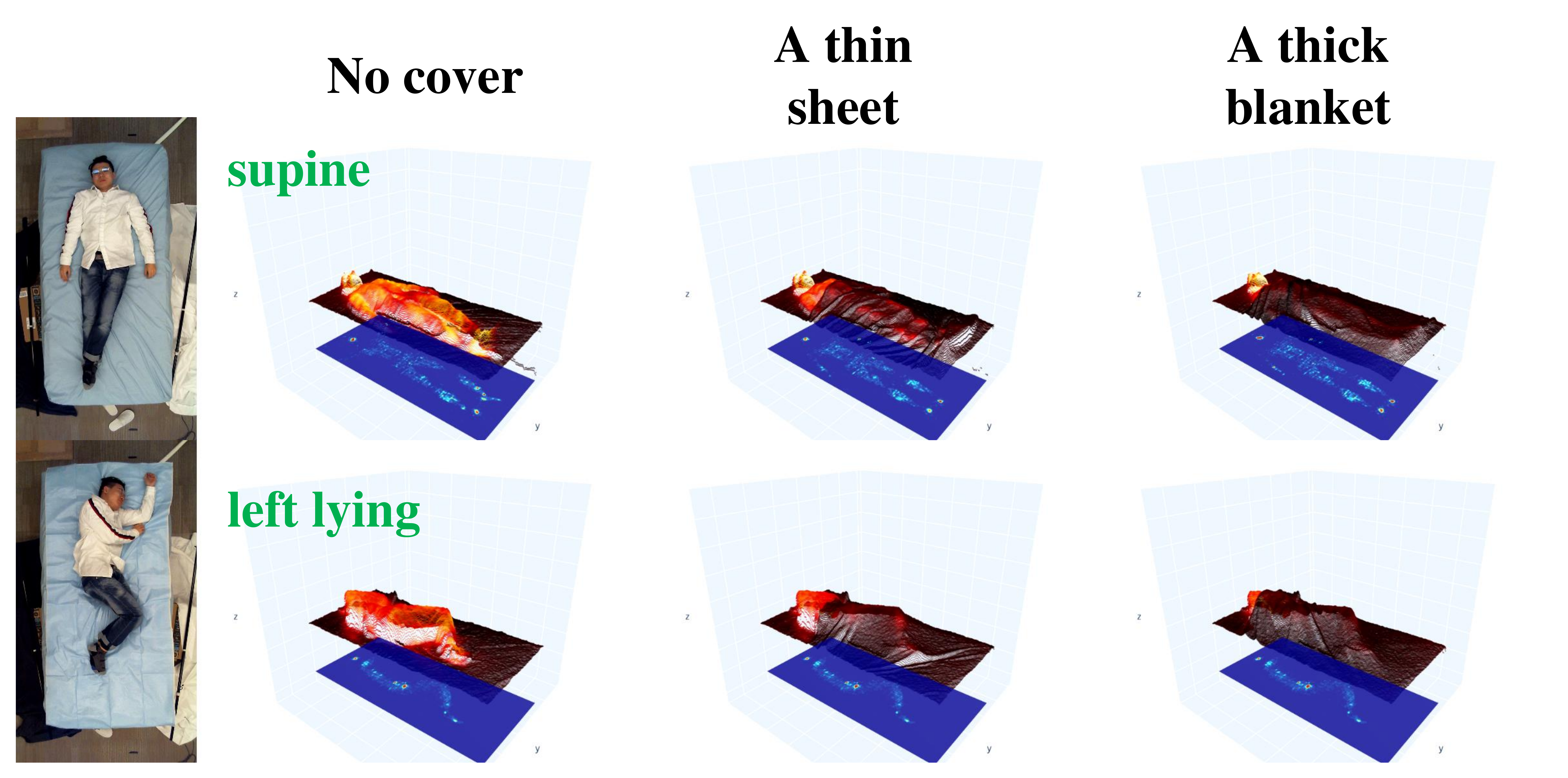}
    \caption{A demo of the LWIR-D-PM visualization tool.}
    \label{fig:fuseVis}
    \vspace{-.25in}
\end{figure}
}
\newcommand{\demoThr}{
\begin{figure}[t]
\vspace{-.11in}
 \centering
 \subfloat{\includegraphics[width=1\linewidth,{trim=0in 0in 0in 0in,
  clip=true}]{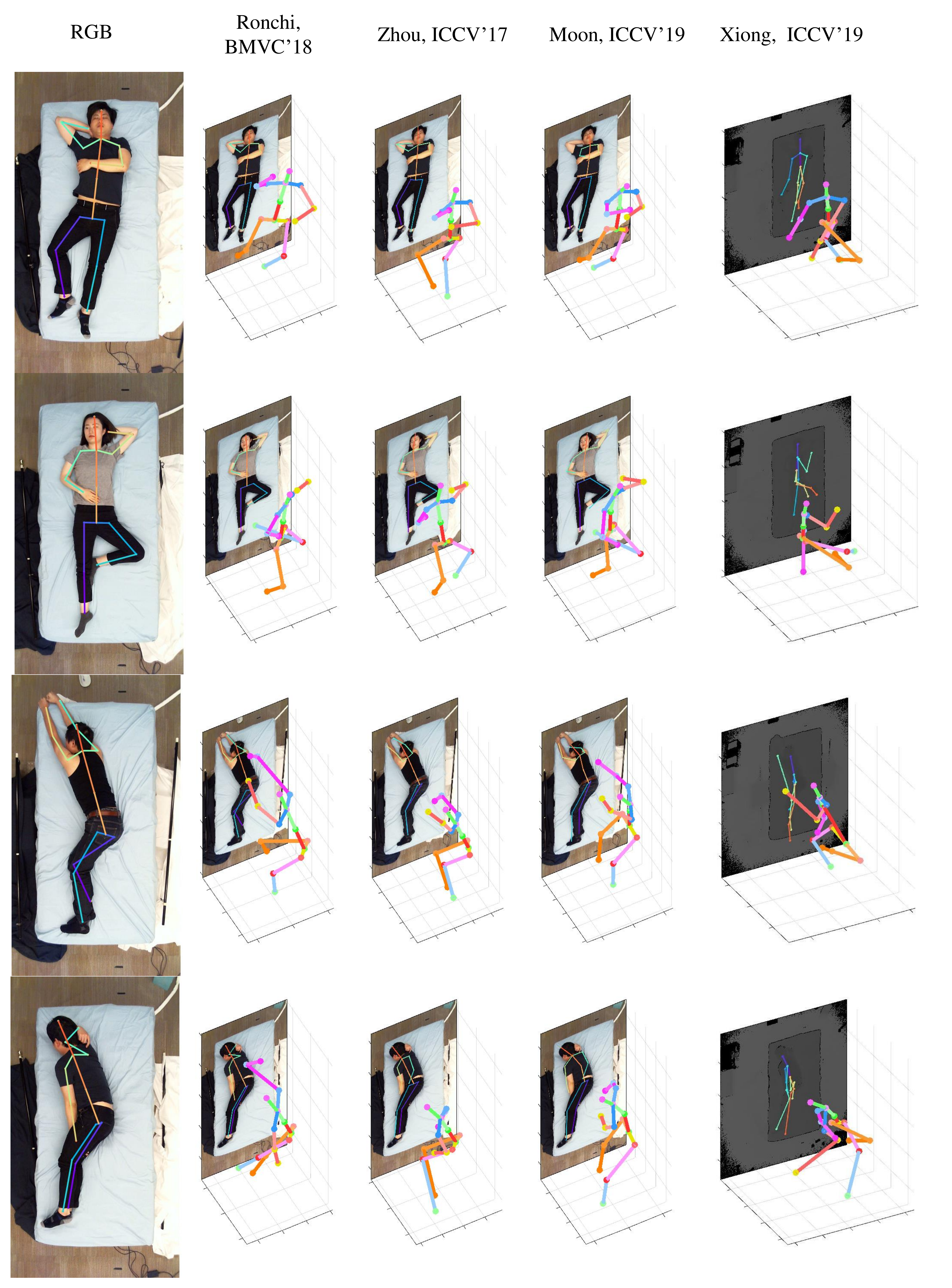}}
   \caption{Qualitative results of 3D in-bed pose estimation. RGB images with ground truth are given followed by 3D pose estimation results from Ronchi et al. (BMVC'17) \cite{relativeposeBMVC18},  Zhou et al. (ICCV'17) \cite{Zhou_2017_ICCV}, Moon et al. (ICCV'19) \cite{Moon_2019_ICCV_3DMPPE}, and Xiong et al. (ICCV'19) \cite{xiong2019a2j}.} 
\label{fig:demoThr}
\vspace{-.2in}
\end{figure}
}

\newcommand{\tblMeanStdThr}{
\begin{table*}[t]
\caption{Mean and variance of two human pose estimation models trained/tested on MPII \cite{andriluka20142d}, COCO \cite{lin2014microsoft}, and SLP datasets. The reported performance is based on the PCKh@0.5 metric.}
\begin{center}
\begin{tabular}{ c  c | c c c c c}
\hline 
Models & Datasets & MPII & COCO & SLP-LWIR & SLP-PM & SLP-Depth  \\
\hline \hline
Sun et al. \cite{sun2019deep} & Mean & 0.167 & 0.167 & 0.172 &  0.214 & 0.157 \\
                            & STD & 0.127 & 0.136 & 0.107 & 0.117 & 0.099 \\
\hline
Xiao et al. \cite{xiao2018simple} & Mean & 0.173 & 0.349 & 0.177 & 0.206 & 0.163 \\
                            & STD & 0.127 & 0.118 & 0.107 & 0.115 & 0.101 \\
\hline
\end{tabular}
\end{center}
\label{tbl:tblMeanStdThr}
\end{table*}
}

\newcommand{\tblSotaMPII}{
\begin{table}[h]
\caption{Mean and STD of the regression errors in normalized pixel for points which are within PCKh@0.5 threshold, when state-of-the-art models are trained on MPII \cite{andriluka20142d} (all in RGB modality)  and SLP (its individual LWIR, PM, and Depth modalities) datasets and tested on corresponding dataset/modality, respectively.}
\begin{center}
\begin{tabular}{ c | *3c || c}
\hline 
 Datasets & \multicolumn{3}{c||}{SLP} & MPII \\
 \hline
 Modalities & LWIR & PM & Depth & RGB \\ 
\hline \hline
Sun et al. \cite{sun2019deep} & 93.4 & 84.3 & 96.4  & 92.3 \\
Xiao et al. \cite{xiao2018simple} & 92.5 & 86.8 & 95.3 & 91.5 \\ 
Gkioxari et al. \cite{gkioxari2016chained} & 88.8 & 88.5 & 94.2 & 85.3 \\
Chu et al.  \cite{chu2017multi} &  89.9 & 88.1 & 94.6 & 91.5  \\ 
Yang et al.  \cite{yang2017learning} &  94.2 & 90.7 & 96.6 & 92.0 \\
Newell et al. \cite{newell2016stacked} & 94.0 & 90.1 & 96.5 & 90.9  \\

\hline
\end{tabular}
\end{center}
\label{tbl:sotaMPII}
\end{table}
}

\begin{document}

\title{Simultaneously-Collected Multimodal Lying Pose
Dataset: Towards In-Bed Human Pose
Monitoring under Adverse Vision Conditions}

\author{Shuangjun Liu, 
        Xiaofei Huang, 
        Nihang Fu,
        Cheng Li,
        Zhongnan Su,
        and~Sarah Ostadabbas$^{*}$
\IEEEcompsocitemizethanks{\IEEEcompsocthanksitem S. Liu, X. Huang, N. Fu, C. Li, Z. Su, and S. Ostadabbas are with the Augmented Cognition Lab, Department of Electrical and Computer Engineering, Northeastern University, Boston,
MA, USA.\protect\\

$^*$Corresponding author's email: ostadabbas@ece.neu.edu.}
}

\markboth{}%
{Shell \MakeLowercase{\textit{et al.}}: Bare Advanced Demo of IEEEtran.cls for IEEE Computer Society Journals}

\IEEEtitleabstractindextext{%
\begin{abstract}
Computer vision (CV) has achieved great success in interpreting semantic meanings from images, 
yet CV algorithms can be brittle for tasks with adverse vision conditions and the ones suffering from data/label pair limitation. One of this tasks is in-bed human pose estimation, which has significant values in many healthcare applications. In-bed pose monitoring in natural settings could involve complete darkness or full occlusion. Furthermore, the lack of publicly available in-bed pose datasets hinders the use of many successful pose estimation algorithms for this task. In this paper, we introduce our Simultaneously-collected multimodal Lying Pose (SLP) dataset, which includes in-bed pose images from 109 participants captured using multiple imaging modalities including RGB, long wave infrared, depth, and pressure map. We also present a physical hyper parameter tuning strategy for ground truth pose label generation under extreme conditions such as lights off and being fully covered by a sheet/blanket. SLP design is compatible with the mainstream human pose datasets,  therefore, the state-of-the-art 2D pose estimation models  can be trained effectively with SLP data with promising performance as high as 95\% at PCKh@0.5 on a single modality. The pose estimation performance can be further improved by including additional modalities through collaboration. 
\end{abstract}

}

\maketitle

\IEEEdisplaynontitleabstractindextext

%
\IEEEpeerreviewmaketitle

\IEEEraisesectionheading{\section{Introduction}\label{sec:introduction}}

\IEEEPARstart{S}leep/at-rest behavior monitoring is a critical aspect in many healthcare prediction, diagnostic, and treatment practices, in which accurately tracking poses that the person takes while in bed plays an important role in the outcomes of the studies in this field \cite{neill1997effects,helm1951importance,ostadabbas2012resource}. These studies reveals that in-bed poses affect the symptoms of many medical complications such as sleep apnea \cite{lee2015changes}, pressure ulcers \cite{black2007national}, and even carpal tunnel syndrome \cite{mccabe2010evaluation}.  
%
%
The need for automatic in-bed behavior monitoring systems  is becoming more apparent especially during the recent pandemic when spiking numbers of patients require consistent monitoring throughout the day \cite{cov19-cdc}. Medical system overload is also commonly observed among the epicenters globally \cite{haleem2020effects}. In such circumstances, automatic patient monitoring systems that can be employed unobtrusively at home or local medical centers, not only could lead to reduced hospital visits and therefore mitigating the risk of infection spread, but also could bring on some workload relief for the already overworked caregivers.   


However until now, in-bed human pose monitoring systems still heavily rely on the obtrusive wearable devices \cite{labuzetta2016using}, or manually-taken  reports from the caregivers \cite{smith1995pressure}. 
On one hand, the expensive medical-grade devices can hardly be offered beyond the professional hospital setting. On the other hand, the behavioral reports are usually subjective and many even contradict among medical wards \cite{ouimet2007incidence}. 

The recent computer vision advancements in the human pose estimation topic have opened up a new avenue for contact-less patient monitoring tasks \cite{yuan2011discriminative,rezaei2019target}.  However, the adverse vision conditions around in-bed human pose estimation such as the extreme illumination changes (including full darkness) and the presence of heavy occlusions (e.g. sheets/blanket) have hindered the state-of-the-art pose estimation algorithm accuracies for in-bed pose cases \cite{liu2019bed}. 

Nonetheless, given the importance of this topic in healthcare applications, in the last decade, consistent effort has been made in order to address the in-bed  pose estimation problem by employing other sensing modalities including pressure mapping systems \cite{liu2013dense,ostadabbas2014bed}, depth sensing \cite{martinez2013bam}, as well as  infrared imaging \cite{liu2019seeing}. Yet, the scale of data in these work are limited by having only a few participants and none of the work has publicly released their datasets to the machine learning/computer vision community. 
Lack of publicly available datasets not only makes it hard to reproduce their results and validate their effectiveness, but also comparison with newly-developed algorithms without a common benchmark has not been possible in this field.

To address the challenges surrounding the development of robust in-bed pose estimation algorithms, we present the \emph{first-ever} large-scale publicly accessible in-bed human pose dataset, called Simultaneously-collected multimodal Lying Pose (SLP).  SLP includes all popular imaging modalities ever used in relevant mainstream in-bed pose estimation studies. In this paper, we focus on introducing  the SLP dataset creation process and its underlying principles, its statistics, demo applications, and performance evaluation of state-of-the-arts  human pose algorithms when are trained/tested on SLP images. We describe SLP data collection paradigm with its detailed technical aspects, which could be helpful for relevant studies in potential applications when multimodal data with correspondence is utilized. This paper aims at serving not only the computer vision community but also the healthcare domain by making the following contributions: 

\begin{itemize}
    \item  Presents a large-scale ($>$100 subjects with nearly 15,000 pose images) human in-bed (i.e. at-rest) pose dataset, SLP, with multiple sensing modalities collected simultaneously including RGB, long wavelength infrared (LWIR), depth (D) and pressure map (PM). SLP can be potentially served as a benchmark for in-bed human behavior analysis studies based on different imaging modalities. 
    
    \item SLP is formed in a compatible way with other mainstream human pose datasets, therefore, state-of-the-art human pose estimation algorithms can effortlessly be trained on it and their performance can be reported in commonly-used pose estimation metrics. 

    \item Addresses the difficulties for pose ground truth generation due to the lack of proper illumination and heavy occlusion by providing practical guidelines based on a novel physical hyperparameter tuning (PHPT) approach and  its underlying reasoning. 

    \item  Presents a novel LWIR-D-PM visualization tool specific for in-bed pose monitoring by fusing multiple modalities which provides an intuitive view for healthcare providers  to investigate physical state of the patient's body during monitoring.  
    
    \item In order to validate SLP dataset diversity and broadness in terms of in-bed poses, besides  evaluating the pose inference models  on our main setting (a regular bedroom as shown in \figref{setup}(a)), we specifically redeployed our system in a simulated hospital room (as shown in \figref{setup}(b)) and collected extra data for this field test. The models trained on the main setting could transfer their learning into the new setting, which proves SLP  versatility\footnote{The code is available at: \href{https://github.com/ostadabbas/SLP-Dataset-and-Code}{\textcolor{cobalt}{github.com/ostadabbas/SLP}}. The SLP dataset can be downloaded at:  \href{https://web.northeastern.edu/ostadabbas/2019/06/27/multimodal-in-bed-pose-estimation/}{\textcolor{cobalt}{SLP Dataset for Multimodal In-Bed Pose Estimation}}.}. 
\end{itemize}

The remainder of this paper is structured as follows. \secref{related} presents an overview of the related work in the area of human pose estimation, in particular in-bed  poses. \secref{SLP} introduces the process of building SLP dataset and generating the ground truth labels for each modality. \secref{eval} describes SLP datasets statistics and provides in-bed pose estimation results using inference models that are trained on SLP. \secref{conclusion} concludes the paper and outlines the future work.


\setup
\section{Related Work}
\label{sec:related}
\textbf{General Human Pose Estimation:} There is a long track record of deep learning based human pose estimation algorithms since the introduction of convolutional pose machine \cite{wei2016convolutional}. 
These algorithms already achieved high performance for 2D human pose estimation \cite{newell2016stacked,cao2018openpose,sun2019deep,rogez2019lcr}, which by now could even be deemed as a solved problem.  As far as 3D human pose estimation, noticeable improvements have also been achieved either by the end-to-end training on real 3D human datasets \cite{h36m_pami} or based on a learned human body template \cite{bogo2016keep}.
Human pose estimation under more general settings has also been addressed such as in the wild  \cite{zhou2017towards} or for multi-person with camera distance awareness \cite{moon2019camera}. 

Though more and more application settings have been explored in general for human pose estimation, only a few of them have focused on when human is lying in a bed. The reason comes in multi-fold. Firstly, the mainstream human pose estimation studies are based on the conventional RGB images which can  hardly be effective under darkness, let alone when human subject is  fully covered. Secondly, even for human annotators, pose ground truth generation under such contexts is very challenging and may not be feasible. Lastly, due to the lack of available large-scale datasets, data-driven approaches can  barely be established.  

\textbf{In-Bed Human Pose Estimation:} RGB data although have been employed for detecting leaving or getting into a bed \cite{ding2009bed}, and general posture estimation \cite{liu2017vision,chen2018patient}, however the study settings in these works are limited with well illuminated environments and having little to no occlusions.  To address the adverse vision conditions in regard to the monitoring of the in-bed  poses, other imaging modalities have been introduced including pressure map, depth data, and our own recent work based on LWIR \cite{liu2019seeing}.

For pose estimation using pressure sensors/mats approaches, authors in \cite{pouyan2013continuous} extracted binary signatures from pressure images obtained from a commercial pressure mat and used a binary pattern matching technique for pose classification. The same group also introduced a Gaussian mixture model (GMM) clustering approach for concurrent pose classification and limb identification using pressure data \cite{Ostadabbas2014}. In parallel, authors in \cite{liu2014bodypart} used pictorial structure model of the body based on both appearance and spatial information  to localize the body parts within pressure images. Moreover, estimating 3D human pose directly from 
pressure map has been explored  recently \cite{casas2019patient,clever20183d,clever2020bodies}. However, the 2D or 3D pose ambiguity issues when body parts lose contact with the pressure sensors have been commonly observed in the relevant studies \cite{clever2020bodies,clever20183d}. Another factor that hinders the mainstream use of the pressure mapping systems is their high cost and difficulty in maintaining/cleaning them, which limits their usage to the professional hospital rooms. 

Depth data has been extensively employed for estimating human poses during rest or sleep due to their invulnerability to the darkness during night time.  
Martinez et al. proposed a bed aligned map (BAM) descriptor based on depth information collected from a Microsoft Kinect camera to monitor the patient's sleeping position (not the full pose) and body movements while in bed \cite{martinez2013bam}. They also
reported the estimation results for simulated covered cases, yet no real human data validation was given.  
Their followup work further added the recognition of high-level activities such as removing bed covers into their framework \cite{martinez2015action}. Yu et al. also  employed the depth data to localize the head and body parts while lying in bed. However their model is limited to  the rough granularity only around torso and head parts \cite{yu2012multiparameter}. 


Other modalities such as near IR \cite{liu2019bed} and LWIR \cite{liu2019seeing} have also been explored for in-bed pose estimation purposes.  However, aside from our previous work \cite{liu2019seeing}, the  datasets for other works are  not publicly available, which makes it hard to reproduce their results, and compare them with each other.  Furthermore, these datasets are usually collected for specific application scenarios  with limited modalities and annotations, which makes comparison across approaches and modalities even harder.  

In the present work, we aim at filling these gaps by publicly releasing an in-bed pose dataset, called SLP that includes simultaneously-collected imaging modalities employed by the state-of-the-art studies for in-bed human pose estimation. SLP dataset provides  accurately labelled ground truth poses for each image even when it is taken  under adverse vision  conditions such as full darkness and/or complete occlusion.  With equivalent magnitude of samples to the well-known general purpose human pose datasets such as LSP \cite{Johnson11} with 12K human image samples, MPII  \cite{andriluka20142d} with 25K samples,  and LIP \cite{liang2018look} with 50K samples, using SLP makes training of the in-bed pose estimation models with deep neural network architecture from scratch 
 possible. With public availability and versatile modalities, SLP can  also be employed as a public benchmark for relevant studies. The multimodal nature of the SLP also allows the cross-domain collaboration and inference possible to overcome the issues specific to a single modality \cite{clever20183d}. 

Our work not only aims at providing the raw materials to the computer vision community under  adverse vision  conditions with multimodal correspondence, but also to the healthcare community who may seek  a functional tool for in-bed human pose monitoring  and actually face similar challenges in practice. Therefore, besides evaluating the effectiveness of our dataset in training 2D and 3D human pose estimation models, we also describe the technical aspects of the SLP dataset forming process in details, in case similar problems need to be solved from scratch in practice. 
Several state-of-the-art models trained on SLP will also be released to provide a handy tool to be employed directly for in-bed human pose monitoring purposes.  

\multimodal

\section{Introducing SLP Dataset}
\label{sec:SLP}
To facilitate the ultimate goal of achieving a robust in-bed  pose monitoring system, in the SLP dataset, we have incorporated high numbers of human subjects in various in-bed poses under extreme conditions such as complete darkness and fully covered cases. SLP dataset therefore has the following characteristics: 

\textbf{(i) Modality coverage:}  Mainstream imaging modalities for in-bed human pose are  covered in the SLP including:  RGB \cite{liu2017vision,chen2018patient}, LWIR \cite{liu2019seeing}, Depth \cite{martinez2013bam} and PM (pressure map) \cite{Ostadabbas2014}.

\textbf{(ii) Different cover conditions:}  Poses are collected under conditions as: no cover, a thin sheet, and a thick blanket.

\textbf{(iii) Scenario coverage:}  The most common application scenarios for in-bed pose estimation task are located in a bedroom or in a hospital room. Besides the main dataset, which is collected under a home setting (from 102 participants), we also collected a specific test set for hospital room (from another 7 new participants) to test generalization of the selected pose estimation algorithms in the field. 

\textbf{(iv) Posture coverage:}  Participants are asked to lie in natural poses evenly among supine, left side, and right side sleep postures categories. For each category, 15 poses are collected under 3 cover conditions using 4 imaging modalities, simultaneously.

\textbf{(v) Additional person-specific measurements:}  To facilitate future in-bed pose/behavior studies,  especially when pressure sensing is involved, we also collected additional person-specific measures including participants' weight, height, gender, and tailor measurements of all their limbs.  

 \textbf{(vi)  A systematic multimodal ground truth generation:}  A physical hyperparameter tuning (PHPT) approach and its underlying reasoning are also presented. 

\subsection{SLP Ground Truth Generation Guidelines}
Aiming at vision-based pose inference under adverse vision conditions (e.g. darkness, occlusion), the inference process is not only challenging for machine but also for human, which makes the ground truth generation difficult. To tackle this challenge, we use a physical hyperparameter tuning (PHPT) concept, first introduced  in our previous work \cite{liu2019seeing}. Here, we recast the concept and explain how our ground truth generation guidelines employ PHPT concept in practice. 

A pose labeling process  can be defined as a function $L$ that maps the image $I_{mod}$ in a  modality $mod \in \{ RGB. LWRI, D, PM \}$ to the target pose state $\beta_t$, as:
\begin{equation}
\label{eqn:infer}
\hat{\beta_t} = L\Big(I_{mod}(\alpha_t, \beta_t, \alpha_c, \beta_c)\Big),
\end{equation}
%
where, $\hat{\beta_t}$ is the estimated target pose, $\alpha_t$ and $\beta_t$ stand for the target appearance and pose, and $\alpha_c$ and $\beta_c$ stand for the context appearance and pose, respectively. 


As mentioned in \cite{liu2019seeing}, pose estimation error $E$  depends on not only the pose terms but also the appearance terms.  As all these parameters (i.e. $\{ \alpha_t, \alpha_c, \beta_c\}$) can be decoupled from $\beta_t$ \cite{liu2018inner}, they can be deemed  as the hyperparameters of the function $L$. Therefore, we can formulate the pose estimation problem  as an  optimization problem:

\begin{equation}
\label{eqn:cost}
\hat{\beta_t}=\displaystyle{\argmin_{\hat{\beta_t}} E(\hat{\beta_t}, \beta_t; \alpha_t, \alpha_c, \beta_c, I_{mod})  }.
\end{equation}

The  estimated target pose, $\hat{\beta_t}$ is conditioned on other terms including $ \alpha_t, \alpha_c, \beta_c, I_{mod}$ during the inference process. For example, human perception can achieve a more accurate $\hat{\beta_t}$ in well-illuminated RGB domain ($I_{RGB}$) with no occlusion ($\beta_c, \alpha_c$: no cover context), which means all these terms can be tuned to improve the inference.  
Unlike commonly referred hyperparameters in mathematical modeling, these variables are directly related to the physical properties of the object, so we call them \textit{physical} hyperparameters. 
Due to the physical constraints, we cannot change them freely, yet we showed that in our application, physical hyperparameters can also be altered effectively to optimize target $L$ performance with prior knowledge.  We employ RGB to LWIR modality mapping as an exemplar to demonstrate the PHPT  guidelines for ground truth generation. As mappings between modalities are similar, these guidelines are generalizable to the other modality pairs. 

\textbf{Guideline I: Labeling with Variable Cover States--}Physical hyperparameters $\{ \beta_t, \alpha_c, \beta_c\}$  cannot be optimized mathematically to their optimal values due to their physical constraints. For example, the subject's appearance can be hardly changed without affecting $\beta_t$. However, altering $\alpha_c$ and $\beta_c$ can be easily achieved without affecting $\beta_t$. So we introduce our first guideline for LWIR image labeling: 

\noindent \textit{\underline{Guideline I:} Perform labeling under settings with the same $\beta_t$ but 
no cover to yield best pose labeling performance.}
 
In practice, we employed Guideline I by collecting data not only from covered individuals but also from the same person without cover for  the same exact pose $\beta_t$. In this way, we actually altered $\alpha_c$ and $\beta_c$. Some examples from such data collection process are shown in different columns of \figref{multimodal}. 
In RGB modality, human pose $\beta_t$ under the cover is hard to be annotated as shown in \figref{multimodal}(c), yet we can get the exact $\beta_t$ via \figref{multimodal}(a). In LWIR modality, when pose $\beta_t$ is  hard to annotate accurately due to its blurriness, we label \figref{multimodal}(g), which supposed to have identical pose as \figref{multimodal}(i).


\figTempAmbi 
\textbf{Guideline II: Cross Domain Referencing--}Though imaging via thermal diffusion is promising for under the cover human monitoring and distinguishing the human from the background \cite{liu2019seeing}, however since body limbs share very similar temperature, they may not be highly distinguishable from each other. Moreover, as human moves in the bed, the ``heat residue'' of the previous pose will result in ghost temperature patterns as the heated area needs time to gradually diffuse heat  (see \figref{tempAmbi}(a)). The heat residue  in real-life in-bed pose monitoring will not be an issue, since the person usually stay in a given pose for a while. However, during SLP data collection,  it led to huge labeling  difficulties due to its misleading effect on pose annotation. Unlike real-life sleeping scenarios, we could not expect the participants stay in a given pose for more than a few minutes, otherwise the data collection duration would have been excessively long. Therefore, we addressed this problem by accelerating the heat diffusion progress via a commercial cooling mat (see \figref{setup}). Nevertheless,  complete elimination of the heat residue in a short time period cannot be  achieved in practice. Another ambiguity in limb localization is when limbs are cuddled together and visual cues in LWIR become misleading. For example, in \figref{tempAmbi}(b), from LWIR modality, it is plausible to assume that the left arm is resting on the torso, however it turned out to be under the head from the RGB reference.

These conditions do not necessarily lead to a fully intractable pose localization problem since there still exist cues in the LWIR image such as the heat residue  having slightly less temperature value compared to the true body location (see \figref{tempAmbi}(a)), or the arm will still show a (although weak) profile beside the head (see \figref{tempAmbi}(b)). Nevertheless, these subtle cues in an LWIR image may be missed by the human annotators, which will not happen if they are given the RGB image counterpart (see RGB images in \figref{tempAmbi}). This motivated us to alter function $I_{mod}$ to improve labeling by introducing second guideline:

\noindent \textit{\underline{Guideline II:} Employ  $I_{RGB}$ counterpart as a heuristic guide to prune out false poses in $I_{LWIR}$.}


\textbf{Guideline III: Cross Domain Labeling with Bounded Error--}Although Guidelines I and II can be employed for most cases to achieve highly accurate ground truth labels, cases exist that a limb is nearly intractable when it is fully overlapped with another body part (e.g., a crossed arm over torso). While in RGB images, the difference between the color/texture of different body parts can be used as differentiating cues, in LWIR images uneven temperature distribution when two limbs cross each other makes the detection of the limbs' boundaries difficult.  This also happens when two limbs share similar heights in the depth modality. Moreover, when clothes have complex wrinkles textures that usually mislead the annotators as fake boundaries. 

Here, using the RGB image counterpart, we expand the Guideline II to include the projection of the $I_{RGB}$'s labels into the $I_{LWIR}$'s labels.
%
%
When two images share the same planar surface, plane to plane mapping is feasible through homography \cite{Hartley2004}. 
However, mapping between two $I_{RGB}$ and $I_{LWIR}$ images taken from a human subject in a given pose will usually result in a ghosting effect in homography mapping \cite{xiang2016image}, which is also well-known in panoramic image  creation \cite{szeliski1997creating}. Since human's top surface while lying in a bed is not a flat plane, when approximated by a plane parallel to the bed, mapping coordinate bias occurs. However,we believe such error is bounded in our settings. 


\figHomoErr
Suppose multiple domain cameras are mounted on the same plane. After homography calibration, one point's coordinates on reference plane can be exactly mapped from one domain to another.  Let $P$ be a point on the reference plane, which has exact mapping from RGB to LWIR domain. If we elevate $P$ by $h$ to a higher position $P'$, the ghosting bias will occur, in which mapped point will not coincide with the corresponding point in another camera, as illustrated in \figref{homoErr}. 
%
%
If we look at $P'$ from RGB camera, its image point in $I_{RGB}$ image is equivalent to its projection $P_{RGB}$, the intersection point of the green solid line and the reference plan. Homography mapping in this case is actually mapping $P'$ projection $P_{RGB}$ to another domain (LWIR). However, from LWIR perspective, the true projection of $P'$ is actually $P_{LWIR}$ instead of $P_{RGB}$, and here the bias occurs.

Suppose the distance from camera mounting plane to the reference plane is $H$; $P'$ is biased from reference plane with $z=h$; distance between two cameras is $d$; and the bias between two projection points in different domain is $b$. From the geometry shown in \figref{homoErr}, we have $b = d \times \frac{h}{H-h}$, where $d$ is a preset value for a specific mounting configuration, however $H$ and $h$ are both variables in real applications. $H$ will be determined by the bed height and also the room ceiling height. $h$ depends on the limb length and how much it sticks out of the bed plane. So the bias $b$ can be hardly determined apriori in practice, however $H$ and $h$ should be bounded due to the practical and physical constraints, in which $h\in[h_{low}, h_{high}]$, and $H\in [H_{low}, H_{high}]$. Namely, a hospital or residence room usually have H about 8--10 feet and a human rest in-bed cannot stick out his limb away from the bed plane more than 3--4 feet when lying.  When $h=0$, then $b=0$ which reduces to the case of exact homography mapping when point is located on the reference plane. When $d=0$, it reduces to the case of pure rotation in which the bias will also be eliminated \cite{szeliski2010computer}. In the worst case, 
a point is biased far away from the reference plane with $h= h_{high}$ and a short ceiling height $H=H_{low}$. So, the bias between mapping of a point from $I_{RGB}$ to $I_{LWIR}$ images is bounded as: 
\begin{equation}
\label{eqn:bound}
    b \in[0, \frac{d\cdot h_{high}}{H_{low}- h_{high}}]
\end{equation}

\skelsPM
Using this bounded mapping error, we propose third guideline for labeling:

\noindent \textit{\underline{Guideline III:} When finding exact joint locations are intractable in one domain, employ labels from another domain with bounded bias via homography mapping.}

This is even more necessary for pressure map (PM) data as pose data in PM are highly ambiguous for human annotators. Most existing PM based pose estimation work can only estimate a rough posture \cite{pouyan2013continuous} or limited numbers of limbs  \cite{Ostadabbas2014} as PM is inherently hard to label. A demonstration of PM ground truth of supine and side (right) lying pose via PHPT is shown in \figref{skelsPM}. In the supine pose example, though annotator can provide plausible pose by linking pressure concentrated area (red dash line), yet we see that 
the support area is not always the joint location. 
In \figref{skelsPM}(a), heel resulted pressure is not necessarily aligned with the limb axis due to the unknown roll motion of the leg. For side lying pose in \figref{skelsPM}(b), it is even worse as we can hardly estimate the pose of the right leg or arms crossed over the chest when they lose contact with the bed surface. However, with the PHPT (guideline III), we achieve correct poses in all cases. Furthermore, compared to the other pose estimation studies using PM, 
joint based pose descriptor provides higher granularity and is compatible to many state-of-the-arts pose estimation models. 
In short, PHPT approach  provides a means to generate more accurate ground truth pose labels even when the labels are inaccessible in a given modality. 

\subsection{Cross Modality Alignment}
Conventionally, a camera model can be calibrated with a checkerboard by estimating its intrinsic and extrinsic parameters \cite{hartley2003multiple}. Between well-calibrated camera systems, one point in one system can be accurately mapped into another if  its depth  is known. However, this approach cannot be used in our SLP dataset for cross modality mapping since: (1) except for the depth modality, depth is  unknown in other SLP imaging modalities; (2) checkerboard will not provide thermal correspondence; and (3) pressure map does not have  a pin hole model unlike other camera-based imaging systems. Instead, since  all SLP modalities are in the form of 2D arrays, we employed homography for cross modality mapping  \cite{hartley2003multiple} with respect to a plane parallel to the bed surface, and shared markers were used  across modalities. 

SLP imaging process involves different modality functions, $I_{mod} \in \{ I_{RGB}, I_{LWIR}, I_{D} , I_{PM} \}$, where each modality responds to specific physical property,   
including visible light reflection in $RGB$,  temperature in $LWIR$, distance in $D$, and pressure in $PM$. For example, $I_{PM}$ only depends on the contact pressure quantities no matter what $RGB$, $LWIR$, and $D$ are. So, we need to use markers that trigger relevant response in each modality. 
Suppose a background image with the modality state  as $[s^0_{RGB}, s^0_{LWIR}, s^0_{D}, s^0_{PM}]$, where  $s^0_{RGB}$ is the bed surface appearance, $s^0_{T}$ is assumed to be the room temperature,  $s^0_{D}$ is the distance from the bed surface to the depth camera, and $s^0_{PM}$ can be deemed as zero since no pressure is applied when bed is not occupied. We designed to alter all modalities jointly via a series of visually-detectable markers to have a new state as $[s^0_{RGB}+s^{\Delta}_{RGB}, s^0_{LWIR}+s^{\Delta}_{LWIR}, s^0_{D}+s^{\Delta}_{D}, s^0_{PM}+s^{\Delta}_{PM}]$, by elevating the temperature (LWIR), height (D), and pressure (PM), and altering the RGB appearance by blocking the background. $[s^{\Delta}_{RGB}, s^{\Delta}_{LWIR}, s^{\Delta}_{D}, s^{\Delta}_{PM}]$  stand for the ``appearance'' shift in different imaging modalities  causes by these shared markers. 

\alignModule
Based on this idea, an alignment markers is designed as shown in \figref{alignModule}(a). It consists of a cylinder jar that can easily be recognized by the RGB modality. A thermal plate is attached on top of the jar powered by the batteries in the chamber to alter the LWIR profile. Added weights inside the jar results in increased pressure profile. The jar height is around 10cm, which also  alters the distance in the depth modality. Due to the possible displacement of pressure sensing mattress during experiment, we recalculated the alignment homography before each session. To facilitate this process and reduce experimenter's workload, we designed an automatic center extraction algorithm by getting the geometric center of each marker's contour. An extraction example in RGB domain is shown in \figref{alignModule}(b). We also developed a manual labeling code as a complementary tool when our algorithm fails to extract the correct centers. Experimenter will judge to use manual tool or not according to the automatic extraction results.

\section{System/Dataset Evaluation}
\label{sec:eval}
\phyDist
\subsection{SLP Dataset Statistics}
SLP dataset includes two subsets, which are collected under different settings. The main set is collected from 102 participants (28 females) in a bedroom environment, which is called ``home setting''. We also collected a smaller dataset from 7 participants (3 females) under a simulated hospital room setting at Northeastern University Health Science Department for field test purposes, which is called ``hospital setting'' (see \figref{setup}). All participants were from Northeastern University student population that responded to our recruitment flyers. Using an institutional review boards (IRB)-approved protocol, we collected pose data from each participant while lying in a bed and randomly changing their poses under three main categories of supine, left side, and right side. For each category, 15 poses are collected. Overall 13,770 image samples for home setting and 945 samples for hospital setting are collected in each of the 4 imaging modalities. Moreover, we changed the cover condition from no cover, to cover one (a thin sheet with $\approx$1mm thickness), and then to cover two (a thick blanket with $\approx$3mm thickness). 

We also collected additional person-specific measurements from each participant including their weight (kg), height (cm),  gender (m/f), as well as tailor measurements (i.e. the circumference) of their bust, waist, hip,  upper/lower arm, thigh, and shank (all in cm). The distribution of this data from all of the participants are shown in \figref{phyDist}



\subsection{SLP Data Collection Procedure}
At the beginning of each data collection session, experimenter was required to instruct the process to the participants, provide them with the IRB approved agreement, and take their tailor measurements for additional references as mentioned above. 
With symmetric assumption, we only measured their right side for paired limbs to simplify the process. 
Cross modality alignment was conducted before the main session. 

To improve efficiency and reduce mistakes during data collection, the whole process was managed by our central control software which dispatch tasks to both  participants and sensor devices to coordinate the human-machine collaboration. Each task is the combination of a pose and a cover condition, which requires a joint operation by sensors and human participants.  
On one hand, logical controller  transformed tasks into audio guides to the experimenter and participant. At the start of each task,  participant was requested to move to another natural pose in the designated posture category and then the experimenter was instructed to alter the cover condition accordingly or relaunch the task in case of false operation. 
On the other hand, logical controller  sent trigger command to synchronized data collection module to drive relevant devices to capture and save data simultaneously to a hard-drive.  


\subsection{Ground Truth Generation via PHPT}
To demonstrate our PHPT guidelines, we will illustrate this process with one modality LWIR for example. 
We labeled the collected LWIR pose images by finding 14 body joints in each, based on three different strategies: (1) LWIR-G1 which employs only Guideline I, (2) LWIR-G3 which employs only Guideline III, and (3) LWIR-G123 which employs all three guidelines. As this is an evaluation of ground truth generation process, and there was no higher level standard to refer to, therefore we used the labeling results of LWIR-G123 as the reference and evaluated how much other strategies are biased from this one using a normalized distance metric (based on the probability of correct keypoints; PCK \cite{andriluka20142d}) to visualize the error distribution when different labelling strategies are used. 

The total differences between the labels from the golden standard (LWIR-G123) and the LWIR-G1 and LWIR-G3 are  shown in \figref{gtDist} as the histograms of normalized distance error with fitted Gaussian curve. Compared to the LWIR-G3, LWIR-G1 error shows lower mean value however larger variance, which demonstrates using LWIR-G1 yields high accuracy for recognizable poses yet has larger error for the ambiguous cases. In contrast, LWIR-G3 causes the ghosting errors that persist throughout the labeling process, but with less significant biases. 

\gtDist

\subsection{In-Bed Pose Estimation Accuracy}
With the similar scale and annotation style of many publicly available human pose datasets,
SLP is compatible for training of  most of the state-of-the-art human pose estimation models, when   their performance can be fairly evaluated with well-recognized metrics employed in CV community. 
To demonstrate this, we trained several pose inference models, including HRnet by Sun et al. (CVPR'19) \cite{sun2019deep}, SimpleBaseLine by Xiao et al, (ECCV'18) \cite{xiao2018simple}, ChainedPredictions by Gkioxari et al. (ECCV'16) \cite{gkioxari2016chained}, PoseAttention by Chu et al. (CVPR'17) \cite{chu2017multi},  PyraNet by Yang et al. (ICCV'17) \cite{yang2017learning}, and StackedHourGlass by Newell et al. (ECCV'16) \cite{newell2016stacked}, from scratch on SLP dataset  and reported their performance based on the PCKh metric in different imaging modalities \cite{andriluka20142d}. 
Due to the adverse vision conditions associated with in-bed pose monitoring, RGB modality would no longer be applicable. So, our evaluation platform is  based on other modalities which are still effective under darkness/occlusion including LWIR, depth, and PM.

\pckDana
\demoTwo

\textbf{Implementation Details--}In each work, we chose one of their typical configurations in our evaluation. In \cite{sun2019deep}, we chose the W32 configuration with width 32 for the high resolution subset.   In \cite{xiao2018simple}, we chose the configuration with RestNet-50 backbone. In \cite{chu2017multi,yang2017learning,newell2016stacked}, we set the stage number as 2. 

All models are adapted to work with the corresponding SLP modalities or joint modalities by varying the models' input channels.  All models are trained from scratch with the corresponding modalities in SLP from the training split which is the first 90 subjects \cite{liu2019seeing}. All models are trained on an NVIDIA V100 GPU with  100 epochs, learning rate 1e-3, Adams optimizer \cite{kingma2014adam}, learning decay rate 0.1 at epoch 70 and 90.  Batch size is set to 30 for \cite{chu2017multi,yang2017learning}, and 60 for other models to fit the GPU memory capacity.  Our augmentation includes rotation, shifting, scaling, color jittering, as well as synthetic occlusion \cite{zhong2017random},  to simulate the potential objects that may block the view-point, such as a bedside table.

\textbf{Evaluations under the Home Setting--}The models' pose estimation performance based on each modality is reported in \figref{pckDana}.  Overall, pose estimation using LWIR and depth  modalities show noticeably higher  performance than PM modality, which complies with the findings in other PM based studies  due to the ambiguity issues when limbs have no  contact with the bed \cite{clever2020bodies,clever20183d}. Depth based inference shows more stable performance compared to LWIR, with all 6 methods having over 90\% at PCKh@0.5 against only 4 methods when LWIR is used.  In our test, \cite{yang2017learning} comes out to have the best performance across all modalities with highest PCKh@0.5 of 94.2\%, 90.7\%, 96.6\% for LWIR, PM, and depth, respectively. 
\cite{newell2016stacked} comes after it with a very similar performance. Some qualitative results from Sun et al. model \cite{sun2019deep}  are shown in \figref{demoTwo}, where we added the RGB pose image counterparts with the ground truth for easy observation, followed by inference result from other modalities.
Furthermore, we  compared these models' performance (in PCKh@0.5 metric) when trained and tested on MPII dataset as a general purpose human pose dataset  \cite{andriluka20142d}, and when trained and tested on SLP dataset, as provided in \tblref{sotaMPII}. Except the PM, all these pose inference models show superior pose estimation performance when trained on SLP images from LWIR and depth modality, compared to the RGB pose images. This observation supports the SLP's capability to train large-scale networks  from scratch. 

\tblSotaMPII

\pckSim
\tblMeanStdThr
\textbf{Field Test under the Hospital Setting--}In the hospital setting, our system was deployed in a simulated hospital room involving  different contexts such as: different ceiling height, different bed (a commercial Hill-Rom hospital bed),  sheets/blankets from different brands/colors, and new participants. This reflects most of the possible changes that could occur when our approach is employed in a real application scenario. We collected pose data from all modalities (except PM) from 7 subjects and tested the performance of our pre-trained pose estimation models against this new dataset. Their results are  shown on \figref{pckSim}. The figure shows that the majority of the models trained on SLP demonstrate a robust performance in this field test and \cite{newell2016stacked} for LWIR  96.5\% and \cite{sun2019deep} for depth   96.1\%, in PCKh@0.5 come out to be the best performers.  In these tests, both \cite{newell2016stacked} and \cite{yang2017learning} show a robust  performance in both the original set test and the field test, where \cite{yang2017learning} is  a revised version of \cite{newell2016stacked} with additional attention mechanism.  




\textbf{Annotation Quality Evaluation--}
In order to evaluate the  annotation quality of our SLP dataset, we looked at some pose inference models' performance when trained/tested on the SLP and compared the pose estimation results when the same models are trained/tested on other large-scale human pose datasets. Accordingly, we employed two recent pose estimation models from Sun et al. \cite{sun2019deep} and Xiao et al. \cite{xiao2018simple}, that are separately trained/tested on two public human pose datasets, MPII \cite{andriluka20142d} and COCO \cite{lin2014microsoft}. We also trained/tested these two models  on different modalities of the SLP dataset. We collected all inference errors of normalized pixel within PCKh@0.5 and reported their mean and standard variance (STD) in \tblref{tblMeanStdThr}.  

The rationale behind using this error metric to evaluate annotation quality is as follows. It is reasonable to assume that the general purpose pose datasets such as MPII and COCO could have more complex pose  and appearance distributions than SLP. When a pose inference model fails in these hard cases, it usually cannot even recognize the rough body part area and the error is usually huge. Therefore, a fair comparison between SLP and other pose datasets should  be without including these hard cases.  By assuming no/few hard cases, all pose inferences within the half of the head size threshold (i.e. PCKh@0.5) are deemed as correct with small enough regression error. This regression error can be decomposed into (1) the error between the estimation and observed annotation, and (2)  the dataset intrinsic error between observed annotation and  its real location. If we assume with no/few hard cases, the error (1) is similar across datasets or is biased by a constant, then the difference of regression error distribution within PCKh@0.5 between datasets is mainly due to the dataset intrinsic error \cite{bishop2006pattern} and the standard deviation (STD) of the error will partially reflect the annotation stability of the dataset.  

\tblref{tblMeanStdThr} demonstrates that 
both models trained on SLP-PM show higher mean error than their counterparts trained on other dataset/modalities (except  model \cite{xiao2018simple} trained on COCO).  This agrees with the localization ambiguity issue of the PM  \cite{clever2020bodies,clever20183d}. Meanwhile,  other datasets/SLP-modalities show similar performance to each other. However, in terms of error STD,  almost all SLP modalities yield lower STD  compared to the MPII and COCO. It partially suggests that our proposed guidelines originated from PHPT can help to produce more reliable and consistent annotation. We believe SLP dataset benefits from the cross-modality referencing which will be  further discussed in our ablation study.

\pckDanaJts
\textbf{Ablation Study--}Different from model-focused works, our ablation study focuses on how SLP modalities influence the pose estimation results by extensive evaluation of individual modalities (LWIR, depth, or PM) and their possible collaborations. For this evaluation,  we chose Sun et al. \cite{sun2019deep} pose estimation model and its estimation results for individual  joints and overall are shown in \figref{pckDanaJts}. According to \figref{pckDanaJts}, in single modality test, LWIR and depth are more effective for pose estimation than PM. However, by collaborating with either LWIR or depth, PM performance can be significantly improved as shown in PM-LWIR and PM-depth subplots. \figref{pckDanaJts} also reveals that PM is not counterproductive for inference and could be complementary to other modalities, as reflected in PM-LWIR and PM-depth subplots, where both show performance improvement over their single modality counterparts. 

The underlying reason could be comprehended  by inspecting the qualitative results in \figref{demoTwo}. For example, the second row of \figref{demoTwo} shows that the PM can hardly estimate the arms that are out of contact with the bed, while LWIR and depth can localize both arms more accurately. On the contrary, in the first row of \figref{demoTwo},  when the head rests on the right arm, the depth modality fails to infer the pose of the arm correctly due to the blocked view, while it is clearly  presented in PM. These examples show the complementary effect of PM on LWIR and depth modalities for pose estimation.  
Furthermore, in PM modality, the better performed joint localization are more likely to happen around supportive area such as hips, shoulders, and heels. 
An interesting fact is that these areas are all high risk areas for developing pressure ulcers or bedsore as discussed in relevant studies \cite{shahin2008pressure}. 

\subsection{LWIR-D-PM Visualization}
Our SLP ground truth generation study revealed that human annotators can hardly recognize the poses that are taken in full darkness or high occlusion, only  based on  the RGB images (look at second and third rows of  \figref{demoTwo}). This would be the same for the healthcare providers in sleep behavior monitoring practices. Furthermore, information such as person's body geometry, its temperature and its contact pressure with the bed cannot be extracted from the RGB modality alone. Therefore, we have developed a multimodal pose visualization tool that combines LWIR, depth and PM modality images and visualizes them simultaneously as shown in \figref{fuseVis}.  As all modalities of SLP are collected with correspondence, this presentation can be effortlessly  generated by rendering the coupling modalities one by one. \figref{fuseVis} shows examples with multiple cover conditions from two general in-bed posture categories. A typical benefit of this LWIR-D-PM visualization could be for relevant  medical studies. For example, to investigate which lying poses will lead to high pressure concentration areas and therefore it is the high risk area for bedsore  development \cite{black2007national}.  

\fuseVis

\subsection{Exploring  3D In-Bed Pose Estimation}
While SLP does not contain the 3D human  pose ground truth information, here we try to investigate if it is possible to estimate 3D poses for in-bed cases  by employing other pre-trained 3D pose estimation models.
We tested several state-of-the-art 3D human pose estimation models based on RGB including Ronchi et al. (BMVC'17) \cite{relativeposeBMVC18},  Zhou et al. (ICCV'17) \cite{Zhou_2017_ICCV}, Moon et al. (ICCV'19) \cite{Moon_2019_ICCV_3DMPPE}, and based on depth including Xiong et al. (ICCV'19) \cite{xiong2019a2j}, by feeding them pose images in the corresponding modalities (only cases with no cover). Some qualitative results are shown in \figref{demoThr}. The results in the last column of \figref{demoThr} show that the depth based pre-trained model of \cite{xiong2019a2j} fails most of the times. 
One reason could be that the depth surfaces of different body parts are blended into each other due to the rest state of in-bed human as shown in \figref{fuseVis}. The other reason could be that the bed surface is tightly attached to the body and is acting as the background. This is not  usually an issue in the
existing depth-based human datasets  such as (ITOP) \cite{haque2016viewpoint}, which is centered around  daily activities. 


\demoThr

On the RGB side, most models  can roughly localize the correct human joints. This complies with our assumption that  in-bed human appearances come from the similar distribution as the ones in general human pose datasets. However, for individual limbs/joints, the estimation inaccuracy exists when carefully look at \figref{demoThr}, especially when the subject rest in an ``in-bed'' specific pose such as resting the head on the arms. These models give varying answers to put the hand forth or back, or completely fail the detection at all as shown in the first row of \figref{demoThr}. 
Another typical issue is the uncertainly of depth of the foot as shown in   the second row of \figref{demoThr}, in which  many models prefer to assume a back stretched leg configuration instead of resting on bed. It is plausible to guess the in-bed human is in standing posture from general human pose dataset perspective. We can imagine that if we see a human raise one leg in a standing posture, we are more likely to assume he is trying to kick something and a back stretched leg will be more reasonable.

\section{Conclusion}
\label{sec:conclusion}
In this work, we introduce the first-ever large-scale in-bed pose dataset, called SLP that includes in-bed pose images simultaneously-collected from four imaging modalities including RGB, depth, long wavelength IR (LWIR), and pressure map. SLP dataset provides  accurately labelled ground truth poses for each image even when it is taken  under adverse vision  conditions such as full darkness and/or complete occlusion.
SLP dataset effectiveness is illustrated in our evaluation experiments when multiple state-of-the-art human pose estimation models exhibited robust performance across different modalities, varying cover conditions,  and home vs.  hospital environments. 


In-bed pose cases are  very rare in the existing pose datasets. The publicly-available human pose datasets such as MPII \cite{sapp2013modec}, COCO \cite{lin2014microsoft}, LSP \cite{andriluka20142d}, and FLIC \cite{johnson2010clustered} are predominantly from scenes such as sports, TV shows, and other daily activities, and none provides any specific in-bed poses. Beside privacy issues which has hampered the large-scale data collection,  in-bed pose images differ from available pose datasets due to the notable differences in lighting conditions throughout a day (with no light during sleep time), people being covered with sheet or blanket during sleep, and also having different pose distribution from the common daily activities \cite{liu2017vision}. So, although SLP dataset is supposed to show an easier pose manifold, it is not necessarily fully covered by the existing pose datasets and more likely it is complementary to them. 

We argue that SLP potential values are not limited to the working examples presented in this paper. In CV field, SLP  presents an exemplar recognition/regress problem under   adverse vision conditions, which can be a good starting point for studying similar problems when RGB is no longer effective. Furthermore, its multimodal nature with correspondence makes SLP dataset a qualified candidate for domain adaptation and transfer learning studies. In healthcare domain,  pre-trained pose estimation models on SLP can provide a handy toolkit to track patient poses while in bed. Reliable yet automatic human pose estimation can provide the foundation for many higher level studies such as patient action recognition or behavior monitoring. We would like to leave these suggestions as open topics for future studies in which SLP can serve the community.

\bibliography{paper}
\bibliographystyle{IEEEtran}

\end{document}